\documentclass{article}

\usepackage[final, nonatbib]{nips_2018}

\usepackage[utf8]{inputenc} 
\usepackage[T1]{fontenc}    
\usepackage{hyperref}       
\usepackage{url}            
\usepackage{booktabs}       
\usepackage{amsfonts}       
\usepackage{nicefrac}       
\usepackage{microtype}      
\usepackage{graphicx}
\usepackage{floatrow}
\usepackage{blindtext}

\newfloatcommand{capbtabbox}{table}[][\FBwidth]

\title{Application of Clinical Concept Embeddings for Heart Failure Prediction in UK EHR data}

\author{
  Spiros Denaxas\\
  Institute of Health Informatics\\
  University College London, UK\\
  \texttt{s.denaxas@ucl.ac.uk} \\
  \And
  Pontus Stenetorp, Sebastian Riedel \\
  Department of Computer Science \\
  University College London, UK \\
  \texttt{\{S.Riedel,P.Stenetorp\}@cs.ucl.ac.uk} \\
 \And
  Maria Pikoula, Richard Dobson, Harry Hemingway \\
  Institute of Health Informatics\\
  University College London, UK\\
  \texttt{\{m.pikoula, r.dobson, h.hemingway\}@ucl.ac.uk} \\
}

\begin{document}

\maketitle

\begin{abstract}
Electronic health records (EHR) are increasingly being used for constructing disease risk prediction models. Feature engineering in EHR data however is challenging due to their highly dimensional and heterogeneous nature. Low-dimensional representations of EHR data can potentially mitigate these challenges. In this paper, we use global vectors (\textit{GloVe}) to learn word embeddings for diagnoses and procedures recorded using 13 million ontology terms across 2.7 million hospitalisations in national UK EHR. We demonstrate the utility of these embeddings by evaluating their performance in identifying patients which are at higher risk of being hospitalized for congestive heart failure. Our findings indicate that embeddings can enable the creation of robust EHR-derived disease risk prediction models and address some the limitations associated with manual clinical feature engineering.
\end{abstract}

\section{Introduction}

Risk prediction models are statistical tools which are used to predict the probability that an individual with a given set of characteristics (e.g. smoking, blood pressure, family history of cancer) will experience a health outcome (e.g. heart attack, type 2 diabetes, death). They are a cornerstone of modern clinical medicine \cite{moons2009prognosis} as they enable clinicians to intervene earlier or chose the optimal therapeutic strategy for a patient. Electronic health records (EHR), data generated during routine patient interactions with healthcare providers \cite{hemingway2017big,jensen2012mining}, offer the opportunity to create risk prediction models in larger sample populations and higher clinical resolution \cite{rapsomaniki2013prognostic} than previously available. Utilizing EHR data however is challenging \cite{morley2014defining,hripcsak2012next,albers2018estimating,paxton2013developing} and a recent review \cite{goldstein2017opportunities} illustrated that EHR-derived predictive models used a median of only
27 clinical features, mostly engineered in a cross-sectional fashion.

 Clinical concept embeddings, i.e. multi-dimensional vector representations of medical concepts, can potentially enable the creation of risk prediction models that make use of a patient's entire EHR (e.g. diagnoses, procedures) and reduce the need for manual feature engineering. Contemporary approaches for learning word embeddings are influenced by the neural language model developed by Bengio et al. \cite{bengio2003neural}. Word embeddings are a very popular way of representing high-dimensional and high-sparsity data in the field of natural language processing and have demonstrated a significant improvement in classification accuracy when combined with existing labelled data \cite{turian2010word} . Popular approaches include \textit{word2vec}\cite{mikolov2013efficient}, which includes the continuous bag of words and skip-gram models, and \textit{GloVe} \cite{pennington2014glove}, which produces word embeddings by fitting a weighted log-linear 
model to aggregated global word-word co-occurrence statistics.


\subsection{Previous research and contribution}

Word embedding approaches have been previously used to create low-dimensional representations of 
heterogeneous clinical concepts (e.g. diagnoses, prescriptions, procedures, laboratory findings) from raw EHR data for various supervised and unsupervised learning tasks \cite{beam2018clinical,shickel2017deep}. Previous research has evaluated the use of 
clinical concept embeddings in US data for predicting the risk of developing heart failure using recurrent \cite{choi2016medical, che2017exploiting} or convolutional \cite{che2017exploiting} neural network architectures. In other disease areas, embeddings have been evaluated for predicting events in critical care patients \cite{farhan2016predictive,johnson2016mimic}, length of stay and associated costs \cite{feng2017patient}, suicide in mental health patients \cite{tran2015learning} and hierarchical regularities in dependencies between health states \cite{miotto2016deep}.

A cornerstone of building risk prediction tools is external replication of findings \cite{collins2014external}. Here, we attempt to replicate and compare, to a certain degree, findings obtained using US data from single hospital care providers \cite{choi2016medical} using UK EHR from multiple providers, and illustrate the application of embeddings for risk prediction. This is the first study, to our knowledge, to utilize UK EHR data in this context. The UK and
US healthcare systems are significantly different in terms of planning, delivery and reimbursement. This in turn directly influences what data are recorded in a patient’s electronic health record. Additionally, in contrast with previous research, we evaluate the predictive performance of different clinical information (e.g. diagnoses, procedures) independently as well since including \textit{all} available data might potentially be counterproductive given the noisy and heterogeneous nature of EHR data.

\section{Methods}

%
%
%
%

\subsection{Data sources and population}

We used secondary care EHR from the UK Biobank \cite{sudlow2015uk}, a population-based research study 
comprising 502,629 individuals in the UK. The study contains extensive phenotypic and genotypic information and longitudinal follow-up for health-related outcomes is through linkages to national EHR from hospital care and mortality registers.   Diagnoses and procedures were recorded using controlled clinical terminologies, i.e. hierarchical ontologies enables clinicians to systematically record information about a patient's health and treatment and enable the subsequent use of data for reimbursement \cite{o2005measuring,bright1989medicaid} and research \cite{bhaskaran2014body,rapsomaniki2014blood}. Diagnoses were recorded using ICD-9 and ICD-10 \cite{world2004international} and procedures using OPCS-4 \cite{de2001survey}. Admitted patients are assigned a primary and up to 15 secondary causes of admission. 

We defined incident and prevalent HF cases using a previously-validated phenotyping algorithm 
from the CALIBER resource \cite{denaxas2012data, koudstaal2017prognostic,gho2018electronic}. HF cases were identified using ICD-9 and 
ICD-10 terms occurring at any position during a patient admission (i.e. primary or 
otherwise) in patients aged 40-85 years old at the time of admission. For patients with multiple HF diagnoses, the date of 
onset was defined as the earliest date of admission. HF cases were matched with four eligible controls on assessment centre, year of recruitment, sex and year of birth. Controls were assigned an index date, which was the date of HF diagnosis of the matched case. We excluded prevalent HF cases from our analyses.

\begin{figure}
\centering
	\includegraphics[scale=0.20]{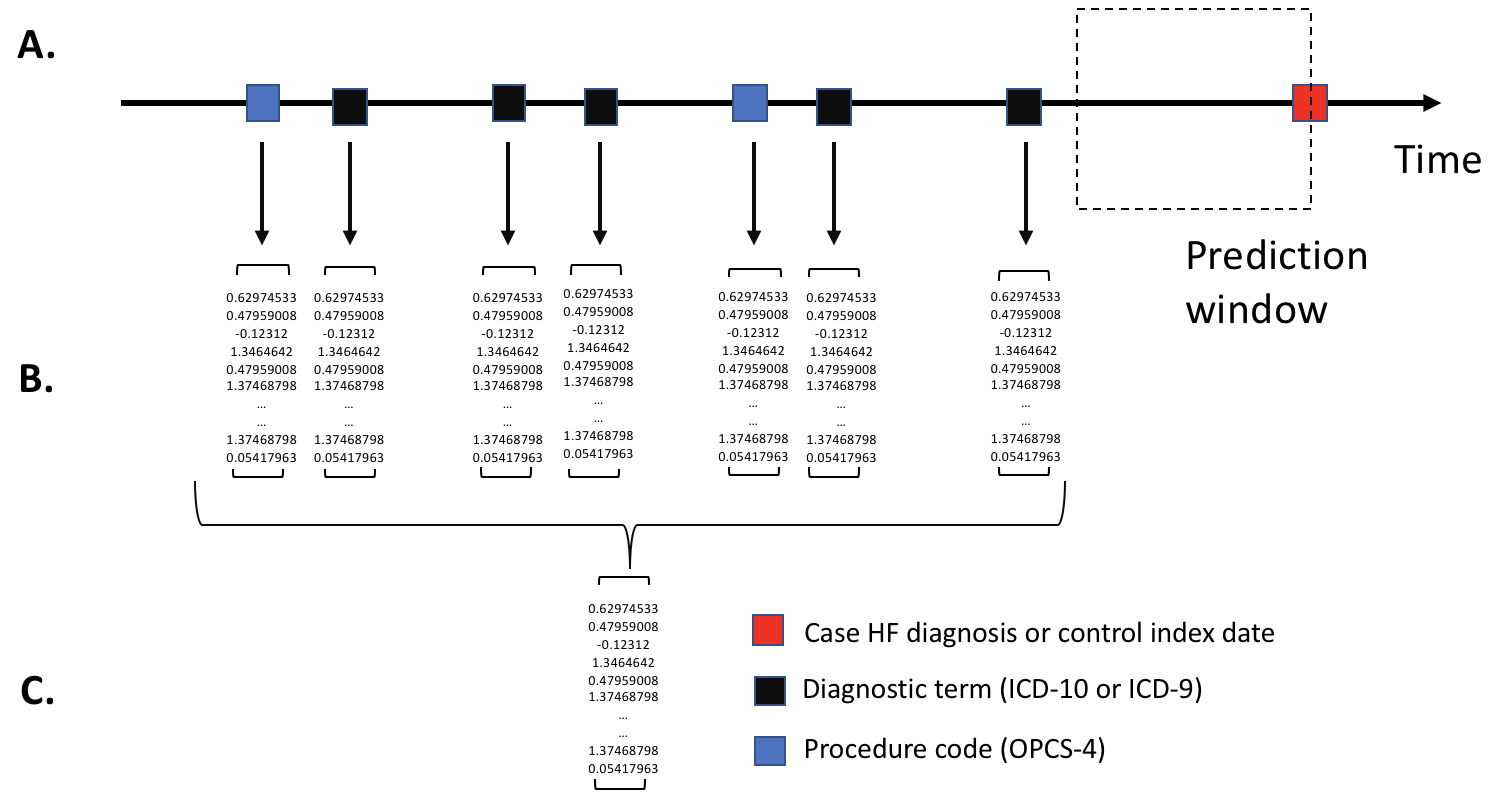}
  \caption{\textbf{(A)} Patient EHR timeline. Concept-level representations of diagnoses and procedures in \textbf{(B)} are transformed into patient-level vector representations. \textbf{(C)} Patients are represented by a vector which is used as input in the supervised risk prediction experiment.}
\end{figure}

\subsection{Learning concept and patient embeddings}
We created four corpuses (Table 1) using: a) 
primary diagnosis terms (PRIMDX), b) primary diagnosis terms and procedure terms (PRIMDX-PROC), c) using primary and secondary diagnosis terms (PRIMDX-SECDX) and, d) primary and secondary diagnosis terms and procedure terms (PRIMDX-SECDX-PROC). We learned \textit{concept-level embeddings} using the \textit{GLoVe} model on the four corpuses and evaluated combinations of embedding dimension (50, 100, 150, 250, 500, 1000) and window sizes (50, 10, 20). Models were trained using Adagrad \cite{duchi2011adaptive} and 150 epochs.
We created \textit{patient-level embeddings} (Figure 1.) by: a) extracting all terms from a patients EHR record from the start of follow up to six months (to exclude features very strongly correlated with a subsequent diagnosis \cite{kourou2015machine}) prior to date of HF diagnoses for cases or the index date for matched controls, b) looking up the vector representations for each embedding, c) creating a vector composed of the mean, max and min of all concept vector representations and, d) normalizing to zero mean and unit variance. For comparisons purposes, we additionally created one-hot representations of EHR data where the feature vector had the same size as the entire vocabulary and only one dimension is on. 

\begin{table}
  \caption{Information on the corpuses used as sources for training the clinical concept embeddings.}
  \label{tcorpus}
  \centering
  \begin{tabular}{lllll}
  Corpus	 & Tokens  & Tokens  & Tokens & Vocabulary  \\
             & (total) & (unique) & (median)	 & size     \\
   \toprule
  PRIMDX             & 2,766,487  & 10,606 & 4  &  5,581  \\
  PRIMDX-SECDX       & 7,699,930  & 13,883 & 7  &  7,797  \\
  PRIMDX-PROC        & 7,904,942  & 18,608 & 11 & 10,949  \\
  PRIMDX-SECDX-PROC  & 12,838,385 & 21,885 & 15 & 13,165  \\
\bottomrule
  \end{tabular}
\end{table}

\subsection{Risk prediction}

We evaluated each set embeddings by applying a linear support vector machine (SVM) classifier to predict HF onset as a supervised binary classification task using the normalized patient-level embeddings as input. We split the data into a training dataset and a test dataset (ratio 3:1) and performed six-fold cross-validation in all modeling iterations on the training data to find the optimal hyper-parameters. We evaluated predictive performance using the area under the weighted receiver operating characteristic curve (AUROC) and the weighted F1 score computed on the test dataset which was unseen. 

\subsection{Implementation}
The SVM was implemented using scikit-learn \cite{pedregosa2011scikit}. \textit{GloVe} embeddings were trained using binaries from pennington2014glove. The documented source code using sample synthetic data for our experiments is available at 
https://github.com/spiros. EHR data used in our experiments are available by applying to the UK Biobank \cite{sudlow2015uk}. UK Biobank ethical approval ref. 9922.

\section{Experimental Results}

We used raw EHR data from 502,639 participants and identified 4,581 HF cases (30.52\% female) 
and matched them as previously described to 13,740 controls. The mean age at HF diagnosis was 
63.397 (95\% CI 63.174-63.619). We trained the clinical concept embeddings from  2,447 
ICD-9, 10,527 ICD-10 and 6,887 OPCS-4 terms across 2,779,598 hospitalizations. We observed that similarly with previous research studies using clinical concept embeddings, diseases which are biologically or contextually closely related across the entire corpus are located close to each other in the vector space (Figure 2).

\begin{figure}
\begin{floatrow}
\ffigbox{%
  	\includegraphics[scale=0.2]{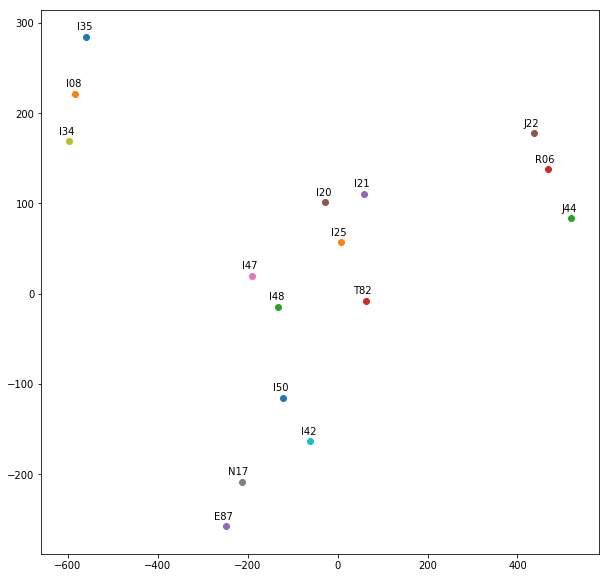}
}{%
  \caption{Ten closest neighbours of the ICD-10 term \textit{I50 Heart Failure} in 2D using t-SNE \cite{maaten2008visualizing}.}%
}
\capbtabbox{%
  \begin{tabular}{llc}
  
    ICD-10 Term & Similarity \\
    \toprule 
    I25 Chronic Ishaemic Heart Disease & 0.521853 \\
    I48 Atrial Fibrillation and Flutter & 0.519000 \\
    R06 Abnormalities in Breathing & 0.479213 \\
    I20 Angina Pectoris & 0.468975 \\
    I21 Acute Myocardial Infarction & 0.460850 \\
    I47 Paroxysmal tachycardia & 0.441402 \\
    N17 Acute renal failure & 0.422741 \\
    I34 Nonrheumatic mitral valve disorders & 0.417978 \\
    I08 Multiple valve disease & 0.412447 \\
    J44 Other COPD & 0.387752 \\
    \bottomrule
  \end{tabular}
}{%
}
\end{floatrow}
\end{figure}

%
%
%
%
%
%
%
%
We observed similar predictive performance across both one-hot and clinical concept embedding prediction experiments. Clinical concept embeddings performed marginally better than one-hot encoded data. The highest performing models were the ones using information combining all diagnoses and surgical procedures (Table 2), obtained with a vector size of 250 and a context window size of five with embeddings. For models using the less extensive corpuses, the best performing results were observed with vectors of smaller size (50 dimensions) and larger context windows (ranging from 10-20). Although, counter-intuitively, the PRIMDX best embedding outperformed PRIMDX-PROC (using procedures and primary diagnoses), PRIMDX-PROC performed better than PRIMDX on average across all vector size and context window combinations. This suggests that clinical concept vectors could be beneficial for risk prediction in absence of a domain ontology or in a semi-supervised fashion combined with labelled data to boost performance \cite{mencia2016medical}. 

\begin{table}[h]
  \caption{Best performing embeddings in test dataset with optimal hyper-parameters.}
  \label{sample-table3}
  \centering
  \begin{tabular}{lllll}
            &   \multicolumn{2}{c}{One-hot} &\multicolumn{2}{c}{Embeddings} \\
  Embedding & AUROC & F1 & AUROC & F1 \\
        
  \toprule 
  PRIMDX             & 0.6543 & 0.7558 & 0.6720  & 0.7390 \\
  PRIMDX-PROC        & 0.6445 & 0.7362 & 0.6662  & 0.7341 \\
  PRIMDX-SECDX       & 0.6697 & 0.7527 & 0.6878  & 0.7568 \\
  PRIMDX-SECDX-PROC  & \textbf{0.6815} & 0.7664 & \textbf{0.6965}  & 0.7500 \\
\bottomrule
  \end{tabular}
\end{table}

Direct comparison with previous studies is challenging due to the use of different 
underlying populations, study designs and incomplete definitions of cohorts and 
outcomes \cite{walsh2014effects,rajkomar2018scalable}. When comparing our results with previous studies 
which used clinical concept embeddings to predict HF onset in a similar experimental setup, our 
approach achieved broadly similar (but slightly worse) overall performance and followed similar patterns: Choi 
\cite{choi2016medical} et al. utilized clinical concept vectors trained using \textit{word2vec} skip-gram and reported an AUROC of 0.711 with one-hot encoded input and AUROC of 0.743 using embeddings with a SVM classifier. Interestingly, the fact that we observed similar (albeit slightly worse) results when using data from multiple hospitals compared to a study sourcing data from a single hospital indicates that embedding approaches can potentially be a very useful tool for scaling analyses across large heterogeneous data source and are insensitive to source variations.
%

\section{Conclusion}
Our work evaluated the use of word embeddings trained using \textit{GloVe} for creating low-dimensionality representations of heterogeneous clinical concepts in UK EHR data. The use of clinical embeddings produced marginally improved predictive performance compared to conventional one-hot models and thus potentially has has numerous applications in healthcare settings where complex, heterogeneous information requires succinct representation or a domain ontology is not fit for purpose. Further research is required to evaluate performance across different prediction windows and increase model interpretability to enable their rapid translation into clinical care.

\newpage

\bibliography{references}

\begin{thebibliography}{10}

\bibitem{moons2009prognosis}
Karel~GM Moons, Patrick Royston, Yvonne Vergouwe, Diederick~E Grobbee, and
  Douglas~G Altman.
\newblock Prognosis and prognostic research: what, why, and how?
\newblock {\em Bmj}, 338:b375, 2009.

\bibitem{hemingway2017big}
Harry Hemingway, Folkert~W Asselbergs, John Danesh, Richard Dobson, Nikolaos
  Maniadakis, Aldo Maggioni, Ghislaine~JM van Thiel, Maureen Cronin, Gunnar
  Brobert, Panos Vardas, et~al.
\newblock Big data from electronic health records for early and late
  translational cardiovascular research: challenges and potential.
\newblock {\em European heart journal}, 39(16):1481--1495, 2017.

\bibitem{jensen2012mining}
Peter~B Jensen, Lars~J Jensen, and S{\o}ren Brunak.
\newblock Mining electronic health records: towards better research
  applications and clinical care.
\newblock {\em Nature Reviews Genetics}, 13(6):395, 2012.

\bibitem{rapsomaniki2013prognostic}
Eleni Rapsomaniki, Anoop Shah, Pablo Perel, Spiros Denaxas, Julie George, Owen
  Nicholas, Ruzan Udumyan, Gene~Solomon Feder, Aroon~D Hingorani, Adam Timmis,
  et~al.
\newblock Prognostic models for stable coronary artery disease based on
  electronic health record cohort of 102 023 patients.
\newblock {\em European heart journal}, 35(13):844--852, 2013.

\bibitem{morley2014defining}
Katherine~I Morley, Joshua Wallace, Spiros~C Denaxas, Ross~J Hunter, Riyaz~S
  Patel, Pablo Perel, Anoop~D Shah, Adam~D Timmis, Richard~J Schilling, and
  Harry Hemingway.
\newblock Defining disease phenotypes using national linked electronic health
  records: a case study of atrial fibrillation.
\newblock {\em PLoS One}, 9(11):e110900, 2014.

\bibitem{hripcsak2012next}
George Hripcsak and David~J Albers.
\newblock Next-generation phenotyping of electronic health records.
\newblock {\em Journal of the American Medical Informatics Association},
  20(1):117--121, 2012.

\bibitem{albers2018estimating}
DJ~Albers, N~Elhadad, J~Claassen, R~Perotte, A~Goldstein, and G~Hripcsak.
\newblock Estimating summary statistics for electronic health record laboratory
  data for use in high-throughput phenotyping algorithms.
\newblock {\em Journal of biomedical informatics}, 78:87--101, 2018.

\bibitem{paxton2013developing}
Chris Paxton, Alexandru Niculescu-Mizil, and Suchi Saria.
\newblock Developing predictive models using electronic medical records:
  challenges and pitfalls.
\newblock In {\em AMIA Annual Symposium Proceedings}, volume 2013, page 1109.
  American Medical Informatics Association, 2013.

\bibitem{goldstein2017opportunities}
Benjamin~A Goldstein, Ann~Marie Navar, Michael~J Pencina, and John Ioannidis.
\newblock Opportunities and challenges in developing risk prediction models
  with electronic health records data: a systematic review.
\newblock {\em Journal of the American Medical Informatics Association},
  24(1):198--208, 2017.

\bibitem{bengio2003neural}
Yoshua Bengio, R{\'e}jean Ducharme, Pascal Vincent, and Christian Jauvin.
\newblock A neural probabilistic language model.
\newblock {\em Journal of machine learning research}, 3(Feb):1137--1155, 2003.

\bibitem{turian2010word}
Joseph Turian, Lev Ratinov, and Yoshua Bengio.
\newblock Word representations: a simple and general method for semi-supervised
  learning.
\newblock In {\em Proceedings of the 48th annual meeting of the association for
  computational linguistics}, pages 384--394. Association for Computational
  Linguistics, 2010.

\bibitem{mikolov2013efficient}
Tomas Mikolov, Kai Chen, Greg Corrado, and Jeffrey Dean.
\newblock Efficient estimation of word representations in vector space.
\newblock {\em arXiv preprint arXiv:1301.3781}, 2013.

\bibitem{pennington2014glove}
Jeffrey Pennington, Richard Socher, and Christopher Manning.
\newblock Glove: Global vectors for word representation.
\newblock In {\em Proceedings of the 2014 conference on empirical methods in
  natural language processing (EMNLP)}, pages 1532--1543, 2014.

\bibitem{beam2018clinical}
Andrew~L Beam, Benjamin Kompa, Inbar Fried, Nathan~P Palmer, Xu~Shi, Tianxi
  Cai, and Isaac~S Kohane.
\newblock Clinical concept embeddings learned from massive sources of medical
  data.
\newblock {\em arXiv preprint arXiv:1804.01486}, 2018.

\bibitem{shickel2017deep}
Benjamin Shickel, Patrick~James Tighe, Azra Bihorac, and Parisa Rashidi.
\newblock Deep ehr: A survey of recent advances in deep learning techniques for
  electronic health record (ehr) analysis.
\newblock {\em IEEE Journal of Biomedical and Health Informatics}, 2017.

\bibitem{choi2016medical}
Edward Choi, Andy Schuetz, Walter~F Stewart, and Jimeng Sun.
\newblock Medical concept representation learning from electronic health
  records and its application on heart failure prediction.
\newblock {\em arXiv preprint arXiv:1602.03686}, 2016.

\bibitem{che2017exploiting}
Zhengping Che, Yu~Cheng, Zhaonan Sun, and Yan Liu.
\newblock Exploiting convolutional neural network for risk prediction with
  medical feature embedding.
\newblock {\em arXiv preprint arXiv:1701.07474}, 2017.

\bibitem{farhan2016predictive}
Wael Farhan, Zhimu Wang, Yingxiang Huang, Shuang Wang, Fei Wang, and Xiaoqian
  Jiang.
\newblock A predictive model for medical events based on contextual embedding
  of temporal sequences.
\newblock {\em JMIR medical informatics}, 4(4), 2016.

\bibitem{johnson2016mimic}
Alistair~EW Johnson, Tom~J Pollard, Lu~Shen, H~Lehman Li-wei, Mengling Feng,
  Mohammad Ghassemi, Benjamin Moody, Peter Szolovits, Leo~Anthony Celi, and
  Roger~G Mark.
\newblock Mimic-iii, a freely accessible critical care database.
\newblock {\em Scientific data}, 3:160035, 2016.

\bibitem{feng2017patient}
Yujuan Feng, Xu~Min, Ning Chen, Hu~Chen, Xiaolei Xie, Haibo Wang, and Ting
  Chen.
\newblock Patient outcome prediction via convolutional neural networks based on
  multi-granularity medical concept embedding.
\newblock In {\em Bioinformatics and Biomedicine (BIBM), 2017 IEEE
  International Conference on}, pages 770--777. IEEE, 2017.

\bibitem{tran2015learning}
Truyen Tran, Tu~Dinh Nguyen, Dinh Phung, and Svetha Venkatesh.
\newblock Learning vector representation of medical objects via emr-driven
  nonnegative restricted boltzmann machines (enrbm).
\newblock {\em Journal of biomedical informatics}, 54:96--105, 2015.

\bibitem{miotto2016deep}
Riccardo Miotto, Li~Li, Brian~A Kidd, and Joel~T Dudley.
\newblock Deep patient: an unsupervised representation to predict the future of
  patients from the electronic health records.
\newblock {\em Scientific reports}, 6:26094, 2016.

\bibitem{collins2014external}
Gary~S Collins, Joris~A de~Groot, Susan Dutton, Omar Omar, Milensu Shanyinde,
  Abdelouahid Tajar, Merryn Voysey, Rose Wharton, Ly-Mee Yu, Karel~G Moons,
  et~al.
\newblock External validation of multivariable prediction models: a systematic
  review of methodological conduct and reporting.
\newblock {\em BMC medical research methodology}, 14(1):40, 2014.

\bibitem{sudlow2015uk}
Cathie Sudlow, John Gallacher, Naomi Allen, Valerie Beral, Paul Burton, John
  Danesh, Paul Downey, Paul Elliott, Jane Green, Martin Landray, et~al.
\newblock Uk biobank: an open access resource for identifying the causes of a
  wide range of complex diseases of middle and old age.
\newblock {\em PLoS medicine}, 12(3):e1001779, 2015.

\bibitem{o2005measuring}
Kimberly~J O'malley, Karon~F Cook, Matt~D Price, Kimberly~Raiford Wildes,
  John~F Hurdle, and Carol~M Ashton.
\newblock Measuring diagnoses: Icd code accuracy.
\newblock {\em Health services research}, 40(5p2):1620--1639, 2005.

\bibitem{bright1989medicaid}
Roselie~A Bright, Jerry Avorn, and Daniel~E Everitt.
\newblock Medicaid data as a resource for epidemiologic studies: strengths and
  limitations.
\newblock {\em Journal of Clinical Epidemiology}, 42(10):937--945, 1989.

\bibitem{bhaskaran2014body}
Krishnan Bhaskaran, Ian Douglas, Harriet Forbes, Isabel dos Santos-Silva,
  David~A Leon, and Liam Smeeth.
\newblock Body-mass index and risk of 22 specific cancers: a population-based
  cohort study of 5{\textperiodcentered} 24 million uk adults.
\newblock {\em The Lancet}, 384(9945):755--765, 2014.

\bibitem{rapsomaniki2014blood}
Eleni Rapsomaniki, Adam Timmis, Julie George, Mar Pujades-Rodriguez, Anoop~D
  Shah, Spiros Denaxas, Ian~R White, Mark~J Caulfield, John~E Deanfield, Liam
  Smeeth, et~al.
\newblock Blood pressure and incidence of twelve cardiovascular diseases:
  lifetime risks, healthy life-years lost, and age-specific associations in
  1{\textperiodcentered} 25 million people.
\newblock {\em The Lancet}, 383(9932):1899--1911, 2014.

\bibitem{world2004international}
World~Health Organization.
\newblock {\em International statistical classification of diseases and related
  health problems}, volume~1.
\newblock World Health Organization, 2004.

\bibitem{de2001survey}
Simon de~Lusignan, Christopher Minmagh, John Kennedy, Marco Zeimet, Hans
  Bommezijn, and John Bryant.
\newblock A survey to identify the clinical coding and classification systems
  currently in use across europe.
\newblock {\em Studies in health technology and informatics}, (1):86--89, 2001.

\bibitem{denaxas2012data}
Spiros~C Denaxas, Julie George, Emily Herrett, Anoop~D Shah, Dipak Kalra,
  Aroon~D Hingorani, Mika Kivimaki, Adam~D Timmis, Liam Smeeth, and Harry
  Hemingway.
\newblock Data resource profile: cardiovascular disease research using linked
  bespoke studies and electronic health records (caliber).
\newblock {\em International journal of epidemiology}, 41(6):1625--1638, 2012.

\bibitem{koudstaal2017prognostic}
Stefan Koudstaal, Mar Pujades-Rodriguez, Spiros Denaxas, Johannes~MIH Gho,
  Anoop~D Shah, Ning Yu, Riyaz~S Patel, Chris~P Gale, Arno~W Hoes, John~G
  Cleland, et~al.
\newblock Prognostic burden of heart failure recorded in primary care, acute
  hospital admissions, or both: a population-based linked electronic health
  record cohort study in 2.1 million people.
\newblock {\em European journal of heart failure}, 19(9):1119--1127, 2017.

\bibitem{gho2018electronic}
Johannes~MIH Gho, Amand~F Schmidt, Laura Pasea, Stefan Koudstaal, Mar
  Pujades-Rodriguez, Spiros Denaxas, Anoop~D Shah, Riyaz~S Patel, Chris~P Gale,
  Arno~W Hoes, et~al.
\newblock An electronic health records cohort study on heart failure following
  myocardial infarction in england: incidence and predictors.
\newblock {\em BMJ open}, 8(3):e018331, 2018.

\bibitem{duchi2011adaptive}
John Duchi, Elad Hazan, and Yoram Singer.
\newblock Adaptive subgradient methods for online learning and stochastic
  optimization.
\newblock {\em Journal of Machine Learning Research}, 12(Jul):2121--2159, 2011.

\bibitem{kourou2015machine}
Konstantina Kourou, Themis~P Exarchos, Konstantinos~P Exarchos, Michalis~V
  Karamouzis, and Dimitrios~I Fotiadis.
\newblock Machine learning applications in cancer prognosis and prediction.
\newblock {\em Computational and structural biotechnology journal}, 13:8--17,
  2015.

\bibitem{pedregosa2011scikit}
Fabian Pedregosa, Ga{\"e}l Varoquaux, Alexandre Gramfort, Vincent Michel,
  Bertrand Thirion, Olivier Grisel, Mathieu Blondel, Peter Prettenhofer, Ron
  Weiss, Vincent Dubourg, et~al.
\newblock Scikit-learn: Machine learning in python.
\newblock {\em Journal of machine learning research}, 12(Oct):2825--2830, 2011.

\bibitem{maaten2008visualizing}
Laurens van~der Maaten and Geoffrey Hinton.
\newblock Visualizing data using t-sne.
\newblock {\em Journal of machine learning research}, 9(Nov):2579--2605, 2008.

\bibitem{mencia2016medical}
Eneldo~Loza Menc{\i}a, Gerard de~Melo, and Jinseok Nam.
\newblock Medical concept embeddings via labeled background corpora.
\newblock In {\em Proceedings of the 10th Language Resources and Evaluation
  Conference (LREC 2016), Paris, France}, 2016.

\bibitem{walsh2014effects}
Colin Walsh and George Hripcsak.
\newblock The effects of data sources, cohort selection, and outcome definition
  on a predictive model of risk of thirty-day hospital readmissions.
\newblock {\em Journal of biomedical informatics}, 52:418--426, 2014.

\bibitem{rajkomar2018scalable}
Alvin Rajkomar, Eyal Oren, Kai Chen, Andrew~M Dai, Nissan Hajaj, Peter~J Liu,
  Xiaobing Liu, Mimi Sun, Patrik Sundberg, Hector Yee, et~al.
\newblock Scalable and accurate deep learning for electronic health records.
\newblock {\em arXiv preprint arXiv:1801.07860}, 2018.

\end{thebibliography}
\bibliographystyle{unsrt}

\end{document}